\pdfoutput=1

\documentclass[11pt]{article}

\usepackage[preprint]{coling}

\usepackage{times}
\usepackage{latexsym}
\usepackage{booktabs}
\usepackage{amssymb}
\usepackage{amsmath}
\usepackage{fixltx2e}
\usepackage[most]{tcolorbox}
\usepackage{listings} 
\usepackage{xcolor} 
\usepackage{listings}
\usepackage{mdframed} 
\definecolor{verylightgray}{gray}{0.97}

\newtcolorbox{promptbox}[1][]{
  colback=verylightgray, 
  colframe=lightgray, 
}
\usepackage[T1]{fontenc}

\usepackage[utf8]{inputenc}
\usepackage{graphicx}
\usepackage{microtype}

\usepackage{inconsolata}
\newenvironment{inconsolatext}{
  \ttfamily 
  \noindent
}{}

\newtcolorbox{roundedprompt}{
  colback=gray!20, 
  colframe=blue!75!black, 
  boxrule=1pt, 
  arc=5mm, 
  boxsep=5pt, 
  left=5pt, 
  right=5pt, 
  top=5pt, 
  bottom=5pt 
}

\title{Context Filtering with Reward Modeling in Question Answering}

\author{Sangryul Kim \\
  KAIST AI \\
  \texttt{sangryul@kaist.ac.kr} \\\And
  James Thorne \\
  KAIST AI \\
  \texttt{thorne@kaist.ac.kr} \\}

\begin{document}
\maketitle
\begin{abstract}
Question Answering (QA) in NLP is the task of finding answers to a query within a relevant context retrieved by a retrieval system. Yet, the mix of relevant and irrelevant information in these contexts can hinder performance enhancements in QA tasks. To address this, we introduce a context filtering approach that removes non-essential details, summarizing crucial content through Reward Modeling. This method emphasizes keeping vital data while omitting the extraneous during summarization model training. We offer a framework for developing efficient QA models by discerning useful information from dataset pairs, bypassing the need for costly human evaluation. Furthermore, we show that our approach can significantly outperform the baseline, as evidenced by a 6.8-fold increase in the EM Per Token (EPT) metric, which we propose as a measure of token efficiency, indicating a notable token-efficiency boost for low-resource settings\footnote{Code and datasets are available at \url{https://github.com/xfactlab/coling2025-context-filtering}}.
\end{abstract}

\section{Introduction}

The ability of language models to effectively understand and process long texts has become a critical requirement, particularly for question-answering (QA) applications \cite{beltagy2020longformer, feldman-el-yaniv-2019-multi, nan-etal-2021-improving, caciularu-etal-2022-long}. However, several studies highlight the problem that even with a substantial amount of relevant context provided, the inclusion of irrelevant content within the context can adversely affect overall performance \cite{10.5555/3618408.3619699, akimoto-etal-2023-context, sauchuk2022role, oh-thorne-2023-detrimental}. The challenge often lies in distinguishing useful information from irrelevant details.

\begin{figure}[t]
    \centering
    \includegraphics[width=1.0\linewidth]{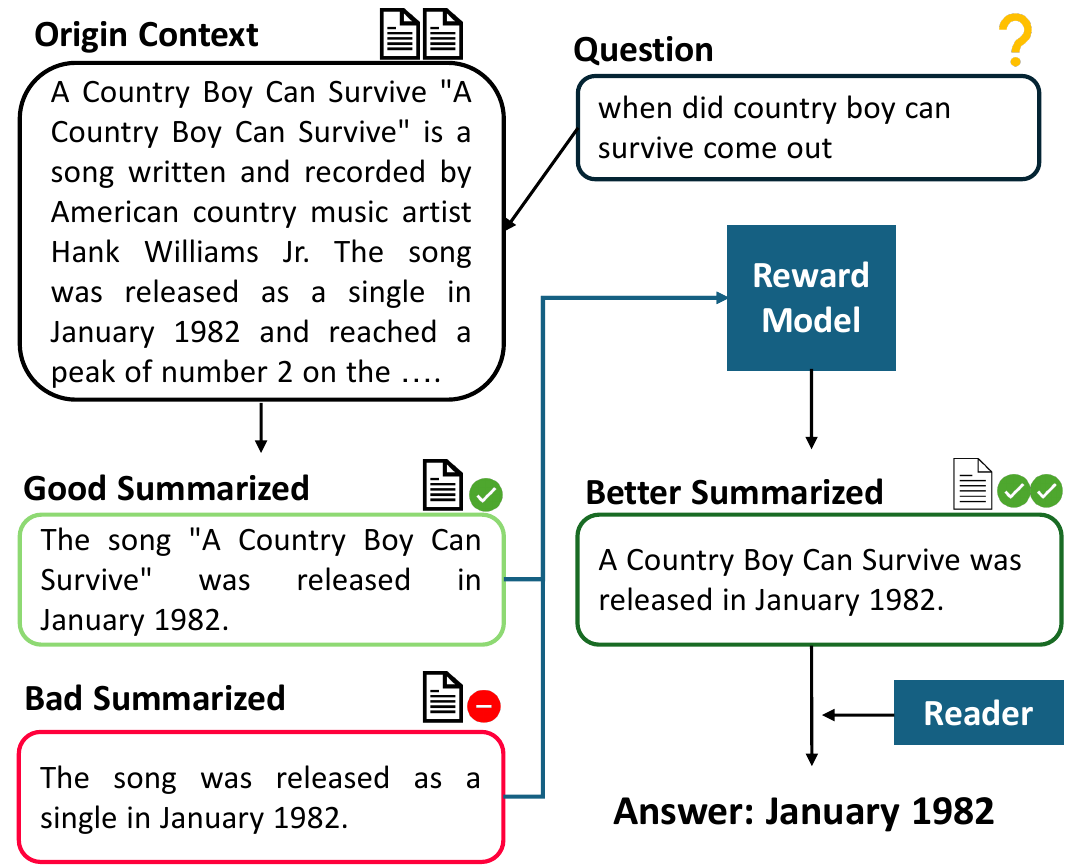}
    \caption{For an effective QA task, we conduct context filtering through the process of creating better summarization using a reward model. Simultaneously, we make it possible to discern which parts are helpful and which are filtered out by utilizing rewards extracted from the data.}
    \label{fig:abstract_img}
\end{figure}

Our research tackles this issue by introducing a new approach to filter out unnecessary content, focusing on summarizing the key points through Reward Modeling. Specifically, we take note of the Direct Preference Optimization (DPO) method \cite{rafailov2023direct}, which trains models using positive and negative feedback from datasets composed of pairs of "\textit{chosen}" and "\textit{rejected}" texts. We suggest employing this technique for filtering the context in QA tasks. We particularly focus on the process of inducing the chosen and negative datasets by paying attention to the presence or absence of specific information within the three essential elements required for the QA task: context, question, and answer. We investigate how the presence or lack of each piece of information impacts the reward modeling process, thus contributing to the development of an efficient context filtering model. Our method ultimately aims to enhance the efficiency of QA models by identifying and retaining only the most relevant information to the query, thereby improving performance. 

We study context efficiency by introducing an EM Per Token (EPT) metric and use it for comparisons between models. This allows us to evaluate the trade-off between context length and the answer Exact Match (EM) score. This is motivated by the fact that more context induces diminishing returns \citep{izacard-grave-2021-leveraging} and comes with performance overheads.
\section{Backgrounds}
\subsection{Knowledge Refinement}
For open-domain question answering, the prevailing trend in research is focused on \textit{retrieving} the correct context from a corpus of knowledge to condition a \textit{reader} \cite{chen-etal-2017-reading, karpukhin-etal-2020-dense, decao2021autoregressive, petroni-etal-2021-kilt}. Simultaneously, language models have been used to augment question information \cite{chen-etal-2023-improving-low} context generation \cite{yu2022generate}, and summarize / paraphrase retrieved information \citep{xu2023recomp, lee-etal-2023-knowledge}. However, recent studies claim that the presence of irrelevant information within the context can lead to a decrease in the performance of the model in a process called \textit{detrimental retrieval} \citep{sauchuk2022role, oh-thorne-2023-detrimental, 10.5555/3618408.3619699, akimoto-etal-2023-context}. Therefore, there is a need for research on models that can filter detrimental content from the retrieved context while aggregating only the essential information.
\subsection{Reward Modeling}
Many methodologies for training LLMs have been developed to align models to user preferences: models are steered away from generating toxic or unhelpful responses. Many methods are derived from the Bradley-Terry model of competition \citep{bradley1952}, using preference pairs containing chosen and rejected responses. In particular, Reinforcement Learning from Human Feedback \citep[RLHF]{ouyang2022training} incorporates human evaluations for training a reward model that is used to score responses, guiding fine-tuning a text generation model to align with user preferences. However, a drawback of this approach is the need to collect human preferences for training, which can be costly. 
Additionally, Proximal Policy Optimization (PPO), commonly used in RLHF, is sensitive to hyperparameter settings. Improper tuning can lead to training instability or divergence \citep{schulman2017proximal, hsu2020revisiting}. Overcoming some of these limitations, \citet{rafailov2023direct} proposed Direct Preference Optimization~(DPO). 
In DPO, the model itself serves as the source of the reward function. 
Let $y_w$ be the preferred output and $y_l$ be the rejected output for a given given task prompt $x$ and denote the dataset $\mathcal{D}=\{x^{(i)}, y_w^{(i)}, y_l^{(i)}\}^N_{i=1}$. We can denote the loss $\mathcal{L}_R(r_\phi, D)$ with the reward function $r_{\phi}(y,x)$ as below \cite{rafailov2023direct}:
\begin{equation}\label{eqn:dpo}
-\mathbb{E}_{(x,y_w,y_l) \sim D} \left[ \log \sigma(r_\phi(x, y_w) - r_\phi(x, y_l)) \right]
\end{equation}
In the QA task, we assume that items with a very high likelihood of containing the answer and whose surrounding context is closely related to the answer are "chosen", while those with a low likelihood of containing the answer and whose surrounding context is likely unrelated to the answer are "rejected". Therefore, we design experiments to find a method that filters only the context necessary for the QA model by conducting training in a way that increases the margin between "chosen" and "rejected" in association with the DPO reward loss.
\section{Context Filtering}

Our experiments are performed studying models for open-domain question-answering. We adopt a \textit{summarize-then-read} pipeline structure adapting the model structure from \citet{inoue-etal-2021-summarize}. The pipeline consists of an abstractive summarization model and a question answering model. 
During the summarize phase, we compare SFT and DPO fine-tuning to determine how efficiently each model summarizes. In the question answering phase, we compare the ability to find the correct answers within the context that has been filtered out through summarization.
We use the FLAN-T5-XL model \cite{chung2022scaling}\footnote{\url{https://huggingface.co/google/flan-t5-xl}} from Hugging Face \cite{wolf-etal-2020-transformers} (3B parameters) throughout all experiments. This seq2seq encoder-decoder model is pre-trained for abstractive summarization and captures the main point of the context. All specific settings, such as prompts and model templates used in the experiments, can be found in Appendix \ref{apdx:propmts}.
\subsection{Data Generation}
\label{sec:data_gen}
Our experiments are conducted on three question answering datasets: SQuAD v1.1 \cite{rajpurkar-etal-2016-squad}, Natural Question (NQ) \cite{kwiatkowski-etal-2019-natural}, TriviaQA (TQA) \cite{joshi-etal-2017-triviaqa}. Retrieved contextual information is selected from DPR \cite{karpukhin-etal-2020-dense}\footnote{\url{https://github.com/facebookresearch/DPR}} for NQ and TQA. For NQ and TQA, instances where the answer to the question is not within the top-1 relevant context are excluded. We split the training set of SQuAD to create an additional validation dataset and use the existing validation set as the test set. The datasets postfixed with "$r$" in the table are those that have undergone this process. Detailed information is in Appendix \ref{apdx:datasets}.

For the QA, the task aims to find the corresponding answer $A$ for a given question $Q$ and a context $C$. We can establish three strategies for summarization through the missing combinations in the necessary information denoted as the tuple $I = (Q, A, C)$. We can construct prompts with different combinations of information: 

\paragraph{Type 1} consists of $I_{1} = (Q, A, C)$, containing all the information (question, answer, and context), aiming to achieve the best possible summarization for comparison and for the highest likelihood of generating correct answers for pairwise training. However, this would not be realistic in an unseen test setting.

\paragraph{Type 2} consists of $I_{2} = (Q, C)$, and the goal is to summarize the parts related to the question in the context without providing information about the answer.  This may be most realistic for application to unseen questions and represents the task signature of the model we would aim to train.

\paragraph{Type 3} is composed of $I_{3} = (A, C)$, focusing on summarizing or extracting related parts in a more lexical aspect, ensuring that the model includes the answer, regardless of the context. We use the outputs generated from the various types of prompt configurations in the following experiments. 


\begin{table*}[h]
\centering
{\small
\begin{tabular}{c|cccc|cccc|cccc}
\toprule
\textbf{Dataset} & \multicolumn{4}{c}{\textbf{NQ$_r$}} & \multicolumn{4}{c}{\textbf{TQA$_r$}} & \multicolumn{4}{c}{\textbf{SQuAD$_r$}} \\ 
\midrule
\textbf{Model} & \textbf{EM} & \textbf{F1} & \textbf{Tok Len} & \textbf{EPT} & \textbf{EM} & \textbf{F1} & \textbf{Tok Len} & \textbf{EPT} & \textbf{EM} & \textbf{F1} & \textbf{Tok Len} & \textbf{EPT} \\ 
\midrule
Origin         & 59.59 & 67.71 & 147.30 & 0.40 & 77.38 & 83.44 & 152.41 & 0.51 & 68.32 & 82.97 & 179.83 & 0.38 \\ 
SFT            & 55.21 & 63.28 & 29.99  & \textbf{1.84} & 69.68 & 75.49 & 28.88  & 2.41 & 59.66 & 76.07 & 30.26  & 1.97 \\ 
DPO$_{O_1, O_2}$ & 49.87 & 58.68 & 41.92  & 1.19 & 66.06 & 72.80 & 39.52  & 1.67 & 48.79 & 61.85 & 35.61  & 1.37 \\ 
DPO$_{O_1, O_3}$ & 52.83 & 59.83 & 29.97  & 1.76 & 62.29 & 68.16 & 18.20  & \textbf{3.42} & 55.40 & 69.28 & 21.46  & \textbf{2.58} \\ 
\bottomrule
\end{tabular}
}
\caption{
Performance comparison across models on different datasets. EM: Exact Match, F1: F1-score, Tok Len: Token Length, EPT: Efficiency per Token.
}
\label{tab:reader_score}
\end{table*}

\subsection{SFT Summarizer Training}
The objective of Supervised Fine Tuning(SFT) is to fine-tune the base model to generate the output summary $O_{1}$ created with the Type 1 prompt when given a context and question (Type 2) as input, learning the policy $\pi^{sft}(O_1 | I_2)$. Through this process, the fine-tuned summarization model ${\pi^{sft}}$ should be able to utilize all the information included in $I_1 = (Q, A, C)$ to produce the best summarization results given $I_2 = (Q,C)$.
\subsection{DPO Summarizer Training}
For the DPO Training, we require pairs of answers $(y_{1}, y_{2})$ and need to determine which summary would be $y_{w}$ and $y_{l}$ satisfying $y_{w}\succ y_{l} \ \vert \ x$ \cite{rafailov2023direct}. Typically human labelers or an LLM determine $y_{w}$ and $y_{l}$, but in this architecture scenario, we assume that outputs from the base model with two types prompts; (Type 2, Type 3) can be candidate of the $y_{l}$, denoted as $O_{2}$, $O_{3}$ respectively. This is because we assume that the outputs from Type 2 and Type 3, which were created with missing information, are less preferred than those from Type 1. We aim to understand how the lack of information in each of Types 2 and 3 affects the reward model through DPO training compared to Type 1. Hense, we construct two different DPO model $\pi^{dpo}_{O_{1},O_{2}}$, $\pi^{dpo}_{O_{1},O_{3}}$. We use Hugging Face TRL  \cite{vonwerra2022trl} for DPO Training.

\subsection{Reader Training}
We study the impact on reducing context through summarization, and thus the reduced context length, on the downstream reader. 
We train the reader to generate answers using the filtered context, which is achieved by summarizing each dataset with the Type 1 trained summarization models. For baseline evaluation, using Type 1 will determine how well each model can deliver answers without loss of information in the "read" phase of the \textit{summarize-then-read} pipeline.

\subsection{Evaluation}
To assess the reader's ability to generate answers using summarized contexts, we use the exact match (EM) score and unigram F1 metrics \cite{rajpurkar-etal-2016-squad}. For the NQ and TQA datasets, where there are multiple possible answers, we initially filtered based on whether the first correct answer was included in the context, treating the first answer as correct if it matched the given answer. Additionally, we calculate the number of tokens required to find the answer in the context using our proposed EM Per Token (EPT) metric. EPT serves as an indicator of the efficiency of the given context, $c$:

\begin{equation}
    EPT(c, y^*, \hat{y}) = \frac{EM(y^*, \hat{y})}{|c|}
\end{equation}

Furthermore, to evaluate the performance of the summarization itself, we introduce a metric called Inclusion Rate of Answer (IRA) shown in Table \ref{tab:ira}. The IRA can verify whether the target answer is fully contained within the given context. If the answer is exactly present, it is evaluated as 1; if it is partially missing or not included, it is evaluated as 0. This metric assesses how well the summarization is done in terms of leaving room for the reader to find the answer, essentially checking the potential for finding the answer before the reader phase. This metric evaluates the summarization process itself, independent of the subsequent reading phase.

\begin{table}[t]
\footnotesize
\centering
\begin{tabular}{cccc}
\toprule
\textbf{Dataset} & $\textbf{NQ}_r$ & $\textbf{TQA}_r$ & $\textbf{SQuAD}_r$ \\ 
\midrule
\textbf{Model} & \multicolumn{3}{c}{\textbf{IRA}} \\
\midrule
Original & 100 & 100 & 100 \\
SFT & 75.93 & 78.76 & 89.94 \\
$\text{DPO}_{O_1, O_2}$ & 71.94 & 77.27 & 66.93 \\
$\text{DPO}_{O_1, O_3}$ & 68.60 & 65.91 & 73.88 \\
\bottomrule
\end{tabular}
\caption{
The ratio indicating whether the target answer is included within the context summarized by each model when the original context is summarized.}
\label{tab:ira}
\end{table}

\section{Impact of Context Filtering}
\paragraph{Trade-off between Token Length and Accuracy Metrics. }

This experiment focuses on extractive QA datasets, where answering a question requires not only identifying the correct answer and its relevant context but also understanding the relationships between different parts of the context that contribute to the answer. By comparing the IRA scores in Table \ref{tab:ira} with the EM scores in Table \ref{tab:reader_score}, we observe that context length does not directly correlate with accuracy. To further investigate, we analyze the contribution of individual tokens to the EM score using the previously defined EPT metric.

Our analysis, conducted with both the SFT and DPO models, demonstrates that models retain accuray with reduced context lengths. For the NQ dataset, reducing the token length to just 20\% of the original retains 92\% of the initial performance. Similarly, for the TQA dataset, a token length of 12\% preserves 80\% of the performance, while for the SQuAD dataset, 12\% of the token length achieves 81\% of the original performance. These results highlight the potential for substantial token reduction without a severe loss in accuracy and underscore potential efficiency gains.


From another perspective, we can observe that the filtering process in SFT and DPO also leads to the loss of information related to the answer span. This can be interpreted in two ways. Firstly, when examining the prompt \textit{"Summarize below context into one sentence..."} used in SFT and DPO, the answer might have been deemed relatively less important in the process of reducing it to one sentence from a summarization standpoint. This suggests that some dependency on the prompt and the model remains, indicating room for improvement. Secondly, our approach to filtering the context did not simply involve lexically cutting off existing content or directly extracting it \cite{wang2023learning}; rather, we restructured it through a prompt, transforming filtering into a summarization task.

Despite this, the results in Table \ref{tab:reader_score} still reveal that the trade-off between token length and EM and F1 metrics is favorable; the efficiency gained from filtering outweighs the slight loss in accuracy, indicating benefits of filtering.

\paragraph{Comparison with Different Reward Model Strategies.}
DPO is known to enable credit assignment down to the token level \cite{rafailov2024from}. Therefore, after training, we can indirectly estimate how it influenced tokens by analyzing the metrics in Table \ref{tab:reader_score} and the EM rates based on length. During the Data Generation phase (Section~\ref{sec:data_gen}), we empirically observe that Type 2 outputs, $O_{2}$, are summarized in a form that provides short answers about the context and question, while the Type 3 outputs, $O_{3}$, are more reflective of the lexical elements surrounding the answer in the context. Furthermore, the presence of the answer in the prompt for Type 3 induces generations that are more similar to Type 1 compared to Type 2. Therefore, during DPO training, the DPO\textsubscript{$O_1, O_2$} model is designed to produce longer outputs that are more similar to those of Type 1, whereas responses from DPO\textsubscript{$O_1, O_3$} are shorter outputs centered around elements present in Type 1 but absent in Type 3 outputs, indicating the intended direction of the training results.
\section{Conclusions}
In this work, we aim to construct an efficient QA system by filtering out unnecessary information from the context. By introducing the EPT metric, we assess the efficiency of the context in QA tasks. The results demonstrate that using the DPO model with reward modeling and its underlying SFT model for filtering is more efficient (per token) than using the full original context. In our future experiments following this study, we plan to apply the reward modeling concept developed here to conduct experiments on effective retrieval models. Moving beyond filtering within a single context, we aim to explore filtering across multiple contexts and incorporate a rewarding model that uses reader performance as a reward. Our goal is to build an integrated Information Retrieval (IR) system that enhances the overall efficiency and effectiveness of the retrieval process.

\section*{Limitations}
In this work, we focus on deriving an efficient context filtering model through reward modeling. The essence of the reward lies in the base model, such as the SFT model evaluating the reward, and the pairs of chosen and rejected for DPO training. We generate data relying on the model's parameters by providing complete information and incomplete information according to each type, without human evaluation, and use this data for our experiments. During this process, we identify the introduction of some unintended biases, leading to biased results and inconsistent outcomes across the datasets. 
Given the nature of the research, it is crucial to generate data that aligns with the intended purpose during the training data generation stage. Therefore, when conducting additional follow-up experiments with this idea, we believe it is necessary to establish proper metrics at the data level and a verification process for evaluating the efficiency and assessing context filtering in QA tasks, separate from the efficiency evaluation.

Moreover, for our experiments, we used datasets that were modified and reduced from existing sources like NQ, TQA, and SQuAD after undergoing specific processing to suit our experimental needs. We believe that adding more diverse datasets, including those covering multi-hop QA \cite{yang-etal-2018-hotpotqa} and long context QA \cite{fan-etal-2019-eli5}, would allow for deeper interpretations.

\section*{Ethics Statement}
In our experiments, we utilize the FLAN-T5 \cite{chung2022scaling} model, which is a T5 \cite{2020t5} model further enhanced with instruction tuning. Therefore, during the training process, the pre-trained parameters could indirectly or directly influence the outcomes, leading to uncontrolled generations that might result in content that is ethically or socially problematic. Consequently, when disclosing such research to the public, it's crucial to be aware of these potential risks and consider implementing engineering and research-based mitigative measures to prevent such issues. We utilize ChatGPT in the process of writing papers, using it for grammar correction and refining sentences. During this process, expressions may be included that differ from the intended meaning.

\section*{Acknowledgments}

This work was partly supported by Institute for Information \& communications Technology Technology Planning \& Evaluation(IITP) grant funded by the Korea government(MSIT) (RS-2019-II190075, Artificial Intelligence Graduate School Support Program(KAIST) and National Research Foundation of Korea(NRF) grant funded by the Korea government(MSIT) (RS-2024-00406715, AI for All: A Social Platform for Barrier-free AI). This work was also supported by the Artificial intelligence convergence cluster development project funded by the Ministry of Science and ICT (MSIT, Korea) \& Gwangju Metropolitan City.

\bibliography{anthology,custom}

\clearpage
\appendix
\section{Hyperparameters}
\label{apdx:hyper}
In our experiments, we conduct three training phases: SFT training, DPO training, and reader training. For each experiment, we utilize one NVIDIA A100 80GB or NVIDIA H100 80GB GPU. We set the number of training epochs to 3, and both the training and evaluation batch sizes to 4, with a gradient accumulation step of 32. The optimizer used is "paged\_adamw\_32bit," with a learning rate of 2e-4, and we employ a "cosine" type learning rate scheduler.

\section{Prompts and Templates}
\label{apdx:propmts}

\subsection{Prompts for Data Geneartion and SFT Training}
\label{apdx:prompts:dg}
\noindent\textbf{[Type 1]}
\begin{inconsolatext}
\begin{promptbox}
\noindent Summarize below context into one sentence according to fit the following context, question and answer.\\
Context: \{context\}\\
Question: \{question\}\\
Answer: \{answer\}\\
Output:
\end{promptbox}
\end{inconsolatext}

\noindent\textbf{[Type 2]}
\begin{inconsolatext}
\begin{promptbox}
Summarize below context into one sentence according to fit the following context and question.\\
Context: \{context\}\\
Question: \{question\}\\
Output:
\end{promptbox}
\end{inconsolatext}

\noindent\textbf{[Type 3]}
\begin{inconsolatext}
\begin{promptbox}
Summarize below context into one sentence according to fit the following context and answer.\\
Context: \{context\}\\
Answer: \{answer\}\\
Output:
\end{promptbox}
\end{inconsolatext}

\subsection{Prompts for Reader Training}
\noindent\textbf{[Train Input]}
\begin{inconsolatext}
\begin{promptbox}
Given the context and question, predict the answer to the question.\\
Context: \{context\}\\
Question: \{question\}\\
Answer:
\end{promptbox}
\end{inconsolatext}

\noindent\textbf{[Train Output]}
\begin{inconsolatext}
\begin{promptbox}
\{target answer\}
\end{promptbox}
\end{inconsolatext}

\section{Details of Datasets}
\label{apdx:datasets}
\begin{table}[h]
\small
\centering
\begin{tabular}{cccc}
\toprule
\textbf{Dataset} & $\textbf{NQ}_r$ & $\textbf{TQA}_r$ & $\textbf{SQuAD}_r$ \\ 
\midrule
\textbf{Split} & \multicolumn{3}{c}{\textbf{Number of Datasets}} \\
\midrule
Train       & 79,618 (43,032) & 78,785 (15,759) & 80,000 \\
Validation  & 8,757 (4,164)   & 8,837 (1,534)   & 7,599 \\
Test        & 3,610 (1,554)   & 11,313 (1,883)  & 10,570 \\
\bottomrule
\end{tabular}
\caption{The specific numbers of data used in the entire pipeline. For NQ and TQA, the numbers in parentheses indicate the actual amount of data involved.}
\label{tab:datasets}
\end{table}

The number of datasets involved in the experiment is depicted in table \ref{tab:datasets}. 

\end{document}